\documentclass[10pt,conference]{IEEEtran}
\usepackage{cite}
\usepackage{amsmath,amssymb,amsfonts}
\usepackage{algorithmic}
\usepackage{graphicx}
\usepackage{textcomp}
\usepackage{xcolor}
\usepackage{url}
\usepackage{hyperref}
\usepackage{float}
\usepackage{stfloats} 
\usepackage{booktabs}
\usepackage{listings}
\usepackage{fancyhdr}

\fancypagestyle{customhead}{
    \fancyhead{} 
    \fancyfoot{} 

    \fancyhead[C]{\textit{\footnotesize This work is under review at International Joint Conference on Neural Networks (IJCNN) at WCCI}}
}

\def\BibTeX{{\rm B\kern-.05em{\sc i\kern-.025em b}\kern-.08em
    T\kern-.1667em\lower.7ex\hbox{E}\kern-.125emX}}
    
\begin{document}

\pagestyle{customhead} 

\title{A Controlled Study of Double DQN and Dueling DQN Under Cross-Environment Transfer}

\author{
    \IEEEauthorblockN{
        Azkaa Nasir\IEEEauthorrefmark{1}, 
        Fatima Dossa\IEEEauthorrefmark{1}, 
        Muhammad Ahmed Atif\IEEEauthorrefmark{1} and 
        Mohammad Shahid Shaikh\IEEEauthorrefmark{1}
    }
    \IEEEauthorblockA{\IEEEauthorrefmark{1}Dhanani School of Science and Engineering, Habib University, Karachi, Pakistan}
}
\maketitle

\thispagestyle{customhead}

\begin{abstract}

Transfer learning in deep reinforcement learning is often motivated by improved stability and reduced training cost, but it can also fail under substantial domain shift. This paper presents a controlled empirical study examining how architectural differences between Double Deep Q-Networks (DDQN) and Dueling DQN influence transfer behavior across environments. Using CartPole as a source task and LunarLander as a structurally distinct target task, we evaluate a fixed layer-wise representation transfer protocol under identical hyperparameters and training conditions, with baseline agents trained from scratch used to contextualize transfer effects. Empirical results show that DDQN consistently avoids negative transfer under the examined setup and maintains learning dynamics comparable to baseline performance in the target environment. In contrast, Dueling DQN consistently exhibits negative transfer under identical conditions, characterized by degraded rewards and unstable optimization behavior. Statistical analysis across multiple random seeds confirms a significant performance gap under transfer. These findings suggest that architectural inductive bias is strongly associated with robustness to cross-environment transfer in value-based deep reinforcement learning under the examined transfer protocol.

\end{abstract}

\section{Introduction}
 
Reinforcement learning (RL) has achieved substantial progress through the integration of deep neural networks, enabling agents to learn complex decision-making policies directly from high-dimensional observations. Value-based methods, particularly the Deep Q-Network (DQN) framework, have played a central role in this progress by extending classical Q-learning to environments with large or high dimensional state spaces. However, standard DQN is known to suffer from several limitations, including overestimation bias and unstable value updates, which can hinder learning performance and generalization.
 
To address these issues, several architectural extensions to DQN have been proposed. Double Deep Q-Networks (DDQN) mitigate overestimation bias by decoupling action selection from action evaluation, resulting in more conservative and stable Q-value estimates. Dueling DQN, on the other hand, introduces a structural decomposition of the action-value function into separate state-value and advantage components, allowing the network to more effectively learn which states are valuable independently of the chosen action. Both architectures have demonstrated improved learning efficiency and stability in single-task benchmark settings.
 
Despite these advances, the extent to which such architectural improvements support transfer learning remains unclear. Transfer learning in RL aims to reuse knowledge acquired in a source task to accelerate or stabilize learning in a target task, thereby reducing sample complexity and training cost. While prior work has explored transfer mechanisms such as policy initialization, representation reuse, and reward shaping, comparatively little attention has been given to how the intrinsic inductive biases of core value-based architectures influence transfer behavior. In particular, it is not well understood whether bias-reduction mechanisms employed by DDQN, or representational decompositions, as used in Dueling DQN, are more conducive to successful transfer under domain shift.
 
This study investigates these questions through a controlled transfer learning setup involving two widely used benchmark environments: CartPole and LunarLander. CartPole is a low-dimensional control task with simple dynamics and dense rewards, making it suitable as a source environment for rapid representation learning. LunarLander, in contrast, presents a more complex control problem with higher-dimensional state representations, sparse and shaped rewards, and long-horizon planning requirements. Transferring policies between these environments therefore constitutes a challenging domain shift, providing a stringent test of representation reuse in value-based RL.
 
The primary focus of this work is the behavior of DDQN and Dueling DQN during transfer to the LunarLander . Performance is evaluated using validation reward as a measure of generalization and robustness. Episode length as a secondary indicator of behavioral efficiency and stability. By fixing the target environment and baseline algorithm, and varying only the architecture of the transferred model, this setup controls for non-architectural factors, allowing architectural effects on transfer learning to be examined. Rather than assuming successful transfer, the analysis emphasizes observed learning dynamics, variance across random seeds, and deviations from a baseline trained from scratch.
 
Accordingly, this study is guided by the following research questions:
 
\begin{itemize}
   \item \textbf{RQ1:} Does transferring a DDQN or Dueling DQN policy from CartPole to LunarLander lead to different validation performance on the target environment?
   \item \textbf{RQ2:} How does the performance of transferred DDQN and Dueling DQN agents compare to a DDQN agent trained from scratch on LunarLander?
   \item \textbf{RQ3:} Do DDQN and Dueling DQN exhibit different stability and behavioral characteristics under transfer, as reflected by variance in validation reward and episode length?
\end{itemize}
 
The contributions of this work are threefold. First, we provide a controlled empirical comparison of DDQN and Dueling DQN under both training from scratch and transfer learning settings. Second, we analyze how architectural inductive bias is reflected in transfer behavior when policies are reused across environments with substantially different dynamics. Third, we highlight stability-related trade-offs and failure modes that emerge during transfer, offering insight into the limitations of naive representation reuse in value-based deep reinforcement learning.
 
By emphasizing comparative behavior and stability rather than absolute performance gains, this work contributes to a more nuanced understanding of transfer learning in deep reinforcement learning. The findings motivate future research into more robust transfer mechanisms and representation learning strategies that can better support generalization across tasks with heterogeneous dynamics.

\section{Related Work and Literature Review}

\subsection{Foundations of Deep Q-Learning and Architectural Extensions}

Despite its success, the baseline DQN architecture suffers from systematic overestimation of action values due to the maximization operation in the Q-learning update rule~\cite{2}. Double DQN (DDQN) addresses this limitation by decoupling action selection from action evaluation, using the online network to select actions and the target network to evaluate them, thereby reducing positive bias in Q-value estimates and improving learning stability~\cite{2,12}. In contrast, the Dueling network architecture proposed by Wang et al.~\cite{15} focuses on representational efficiency by decomposing the Q-function into separate state-value $V(s)$ and action-advantage $A(s,a)$ streams, enabling more effective value estimation when actions have similar returns. These architectural refinements are complementary in nature and have been combined in the Rainbow framework, which integrates Double Q-learning, Dueling networks, and other extensions to achieve strong single-task performance on Atari benchmarks~\cite{18}.

\subsection{Transfer Learning in Deep Reinforcement Learning}

Transfer learning addresses an important challenge in deep reinforcement learning, namely sample inefficiency. Taylor and Stone~\cite{4} provide the foundational framework for transfer learning in RL, defining it as using pre-existing knowledge from one or more source tasks to accelerate learning in a target task. Their structure distinguishes between inter-task mappings, representation transfer, and policy reuse, establishing that successful transfer depends mainly upon task similarity and the transferability of learned representations~\cite{4}.

In deep RL, Zhu et al.~\cite{19} present a detailed and thorough survey categorizing transfer methods by the type of knowledge transferred i.e., reward shaping, policy transfer, inter-task mappings, and representation transfer. Their analysis reveals that while transfer learning can substantially reduce the sample complexity, its effectiveness varies significantly with respect to architectural choices and the extent of domain shift between source and target environments~\cite{19}. Notably, the survey identifies a gap in understanding how fundamental architectural differences in value-based methods such as those between DDQN and Dueling DQN affect transfer learning efficiency.

Progressive neural networks~\cite{8} utilize lateral connections to prevent forgetting and enable positive transfer across Atari games~\cite{8}, while meta-learning approaches such as MAML~\cite{9} optimize for rapid adaptation by learning the initialization parameters that can be transferred~\cite{9}.

Studies provide practical insights into value-based transfer learning. Saadat and Zhao~\cite{17} demonstrate that transferring frozen early layers from single-player to multi-agent environments reduces training time significantly, suggesting that low-level features generalize across related tasks, a principle observed in computer vision transfer learning~\cite{13}. Additionally, Q-value stability emerges as critical for successful transfer: Mishra and Arora~\cite{16} show that Huber loss integration produces smoother learning curves, while Elsaid and Aly~\cite{12} confirm that DDQN's bias reduction mechanisms enable more consistent policy convergence across hyperparameter configurations.

\subsection{Comparative Studies and Identified Gaps}
Recent comparative studies highlight the performance trade-offs among DQN variants across different environments. In a systematic evaluation of DQN, DDQN, DDPG, and PPO~\cite{10} on \texttt{LunarLander-v2}, Elsaid and Aly~\cite{12} find that, although DDQN offers better stability and less overestimation than vanilla DQN, policy-gradient techniques like PPO achieve higher absolute performance. Their analysis highlights how sensitive value-based approaches are to hyperparameters, specifically learning rate ($10^{-4}$ to $5 \times 10^{-4}$), discount factor $\gamma = 0.99$, and epsilon decay rates~\cite{12}. Arulkumaran et al.~\cite{6} note that Dueling architectures tend to excel in high-dimensional state spaces with sparse rewards, while DDQN's bias reduction proves most beneficial in environments with noisy reward signals~\cite{6}.

The challenge of reproducibility in deep RL, highlighted by Henderson et al.~\cite{7}, underscores the importance of rigorous experimental methodology, as random seed selection and network initialization can produce variance comparable to algorithmic differences~\cite{7}. Kancharla~\cite{14} demonstrates the substantial impact of replay buffer size, batch size, and update frequency on DQN convergence in \texttt{LunarLander}, informing our hyperparameter selection~\cite{14}.

Despite extensive research on DQN architectural variants and transfer learning methodologies, no prior work has directly compared how DDQN and Dueling DQN differ in their intrinsic transferability across environments with distinct dynamics. Existing studies evaluate these architectures primarily in single-task settings~\cite{1,2,15} or examine transfer learning using different base algorithms~\cite{17,19}. The literature establishes that both DDQN and Dueling DQN improve upon vanilla DQN through different mechanisms—bias reduction versus value decomposition—but their comparative effectiveness for knowledge transfer remains unexplored.

This study addresses this gap through a controlled empirical investigation of transfer learning from \texttt{CartPole-v2} to \texttt{LunarLander-v3}. By isolating architectural effects through standardized hyperparameters~\cite{12,14}, consistent transfer methodology, multiple experimental runs~\cite{7}, and comprehensive evaluation metrics, we determine which architectural inductive bias better supports generalization across tasks. This investigation extends Taylor and Stone's framework~\cite{4} and Rainbow's insights~\cite{18}, contributing actionable knowledge for practitioners designing transfer learning systems in value-based deep RL, particularly for real world applications where pre-training on simpler tasks can reduce sample complexity in resource-constrained environments~\cite{6,19}. In contrast to prior work that primarily compares transfer mechanisms or algorithms, this study fixes the transfer method and examines how architectural inductive bias alone shapes transfer behavior.

\section{Experimental Setup}

This study investigates the comparative transfer-learning performance of Double Deep Q-Network (DDQN) and Dueling DQN architectures across two reinforcement learning environments: \texttt{CartPole-v2} and \texttt{LunarLander-v3} from the OpenAI Gymnasium suite. Both environments provide control tasks with discrete action spaces but varying state space complexity, offering a controlled setting to analyze each algorithm's efficiency, stability, and adaptability. All hyperparameter values were selected based on established best practices from prior literature~\cite{2,11,12,14,15}, ensuring consistency with standard configurations and allowing architectural differences to be isolated from confounding hyperparameter effects. The methodological framework is divided into five main stages: (1) experimental setup, (2) baseline training, (3) transfer learning implementation, (4) evaluation metrics, and (5) implementation dependencies.

\subsection{Environments and Architecture}

We evaluate DDQN~\cite{2} and Dueling DQN~\cite{15} in both single-task and transfer learning settings using \texttt{CartPole-v2} and \texttt{LunarLander-v3} from OpenAI Gymnasium. CartPole serves as the source domain with its 4-dimensional state space and simple balancing dynamics, enabling rapid convergence~\cite{11,14}. LunarLander provides the target domain with 8-dimensional continuous state space, sparse rewards, and complex multi-dimensional thrust control~\cite{12}. This environment pairing tests cross-domain generalization from simple to complex control tasks~\cite{4,19}.

Agents learn by observing state $s_t$, selecting action $a_t$ via $\epsilon$-greedy policy, receiving reward $r_t$, and transitioning to $s_{t+1}$, with the objective of maximizing cumulative discounted reward $R_t = \sum_{k=0}^{\infty}\gamma^k r_{t+k+1}$. Q-values are updated using temporal-difference learning~\cite{3}:
\[
Q(s_t,a_t) \leftarrow Q(s_t,a_t) + \alpha \left[ r_t + \gamma \max_{a'} Q(s_{t+1},a') - Q(s_t,a_t) \right]
\]

DDQN decouples action selection and evaluation by using the online network to select the greedy action and the target network to evaluate it, reducing overestimation bias from the maximization operation~\cite{2,12}.Dueling DQN splits the shared feature encoder into two streams: one estimating state value $V(s)$ and another computing action advantages $A(s,a)$, which are combined using mean subtraction aggregation. This decomposition enables better generalization when action choice minimally affects long-term returns~\cite{15,18}.

Both architectures use three fully connected layers (128, 128, 64 units) with ReLU activations. For Dueling DQN, the final shared layer splits into separate value and advantage streams before recombining. All networks use experience replay with buffer size 100,000 and batch size 64 to decorrelate training samples~\cite{1,18}.

\subsection{Training Configuration and Hyperparameters}

Each architecture was trained for 500 episodes per environment with discount factor $\gamma = 0.99$~\cite{12,14}. Learning rates were set to $\alpha = 0.0001$ for CartPole and $\alpha = 0.0005$ for LunarLander to balance stability and convergence speed~\cite{11,14}. Epsilon-greedy exploration started at $\epsilon = 1.0$, decaying to $\epsilon_{\min} = 0.01$ at rate 0.995 to ensure appropriate exploration-exploitation balance~\cite{11,12}. Target networks were updated using soft updates with parameters $\tau = 0.005$ (CartPole) and $\tau = 0.001$ (LunarLander)~\cite{1,2}. The Adam optimizer was used with gradient clipping at maximum norm 10 to prevent exploding updates. Following Henderson et al.~\cite{7}, all experiments used multiple random seeds (five seeds) to ensure reproducibility and account for stochasticity in deep RL training.

\subsection{Transfer Learning Protocol}

After baseline training on CartPole (convergence typically achieved at 150-200 episodes with smoothed reward $\geq 195$), we applied transfer learning by reusing the two hidden fully connected layers (128, 128 units) from the pre-trained models. Given CartPole’s low-dimensional state space, we hypothesize that higher-level abstractions rather than raw state encoders are more likely to capture transferable structure. These layers are hypothesized to encode higher-level state-action abstractions to contain transferable knowledge applicable to LunarLander~\cite{4,19}. The input and output layers were reinitialized to match LunarLander's 8-dimensional state space and 4-action output. This representation-transfer approach~\cite{19} aligns with empirical findings by Saadat and Zhao~\cite{17} that freezing early layers preserves generalizable features.

The transferred layers remained frozen for the first 100 episodes of LunarLander training to preserve source-domain knowledge, then were unfrozen for fine-tuning under identical hyperparametersThe freezing period was chosen to preserve source domain representations during early adaptation. Baseline agents trained from scratch on LunarLander provided comparison points to quantify transfer benefits in terms of convergence speed and final performance.

We emphasize that this transfer protocol is intentionally simple and is not claimed to be optimal. The objective is not to maximize transfer performance, but to stress-test architectural robustness under naive representation reuse. By fixing the transfer method and minimizing additional adaptation mechanisms, this design isolates architectural effects and allows negative transfer to emerge clearly when present.

\subsection{Evaluation Metrics}

Performance was assessed using three standard metrics from deep RL research~\cite{2,12,15,18}: (1) \textit{episode reward}—cumulative reward per episode reflecting immediate learning progress; (2) \textit{validation reward}—greedy policy evaluation ($\epsilon = 0$) performed every 10 episodes to assess generalization stability; (3) \textit{training loss}—mean-squared temporal-difference error monitoring Q-value estimation quality and detecting divergence. Success thresholds were defined as reward $\geq 195$ for CartPole and $\geq 200$ for LunarLander~\cite{11,12}. All quantitative results were averaged over multiple random seeds to ensure statistical reliability~\cite{7}.

\section{Results}

\begin{figure*}[t]
\centering
\begin{minipage}{0.32\textwidth}
\centering
\includegraphics[width=\textwidth]{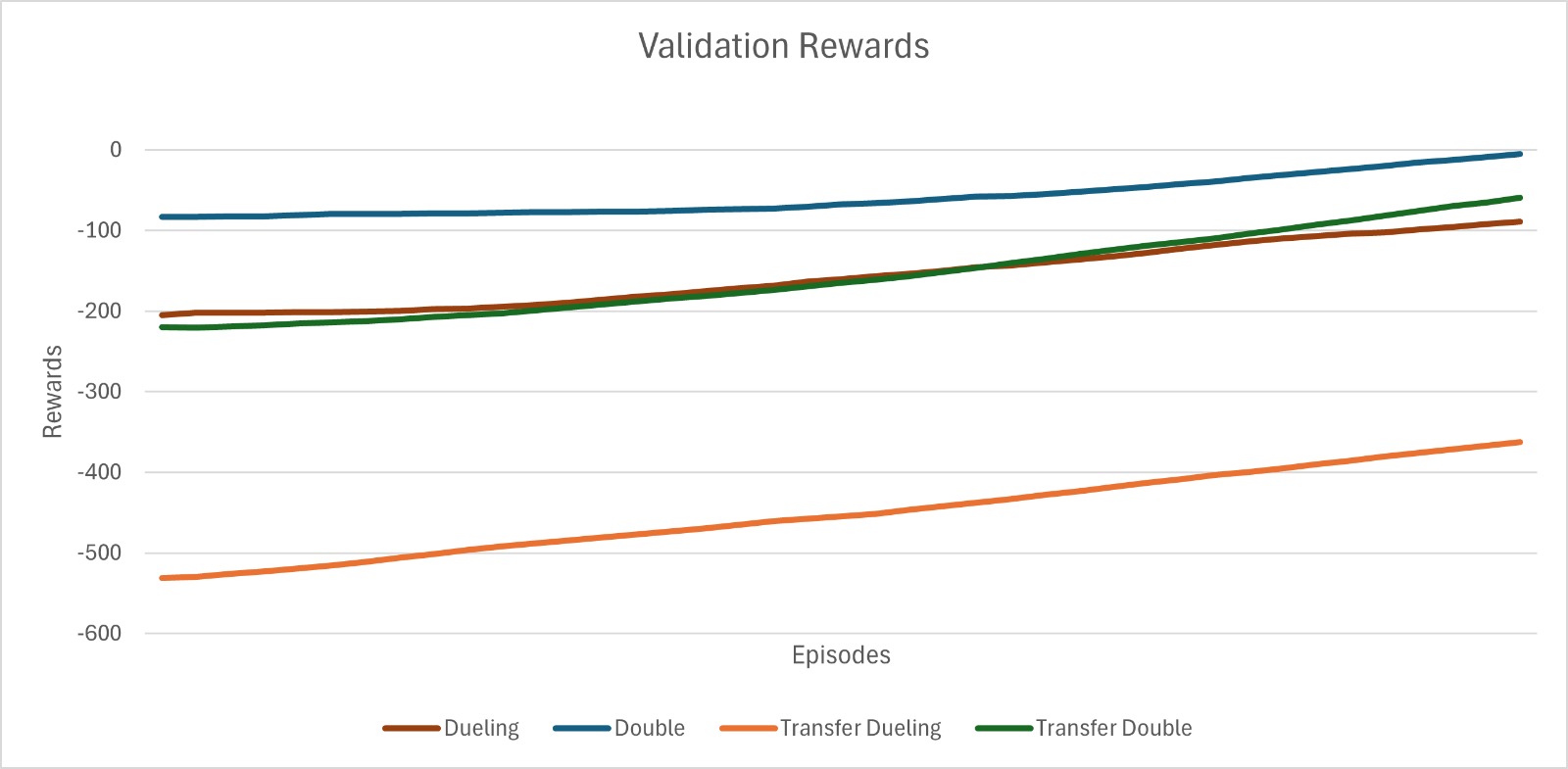}
\caption{Validation rewards: Transfer DDQN stable (-60) vs. Transfer Dueling degraded (-370).}
\label{fig:val_rewards}
\end{minipage}
\hfill
\begin{minipage}{0.32\textwidth}
\centering
\includegraphics[width=\textwidth]{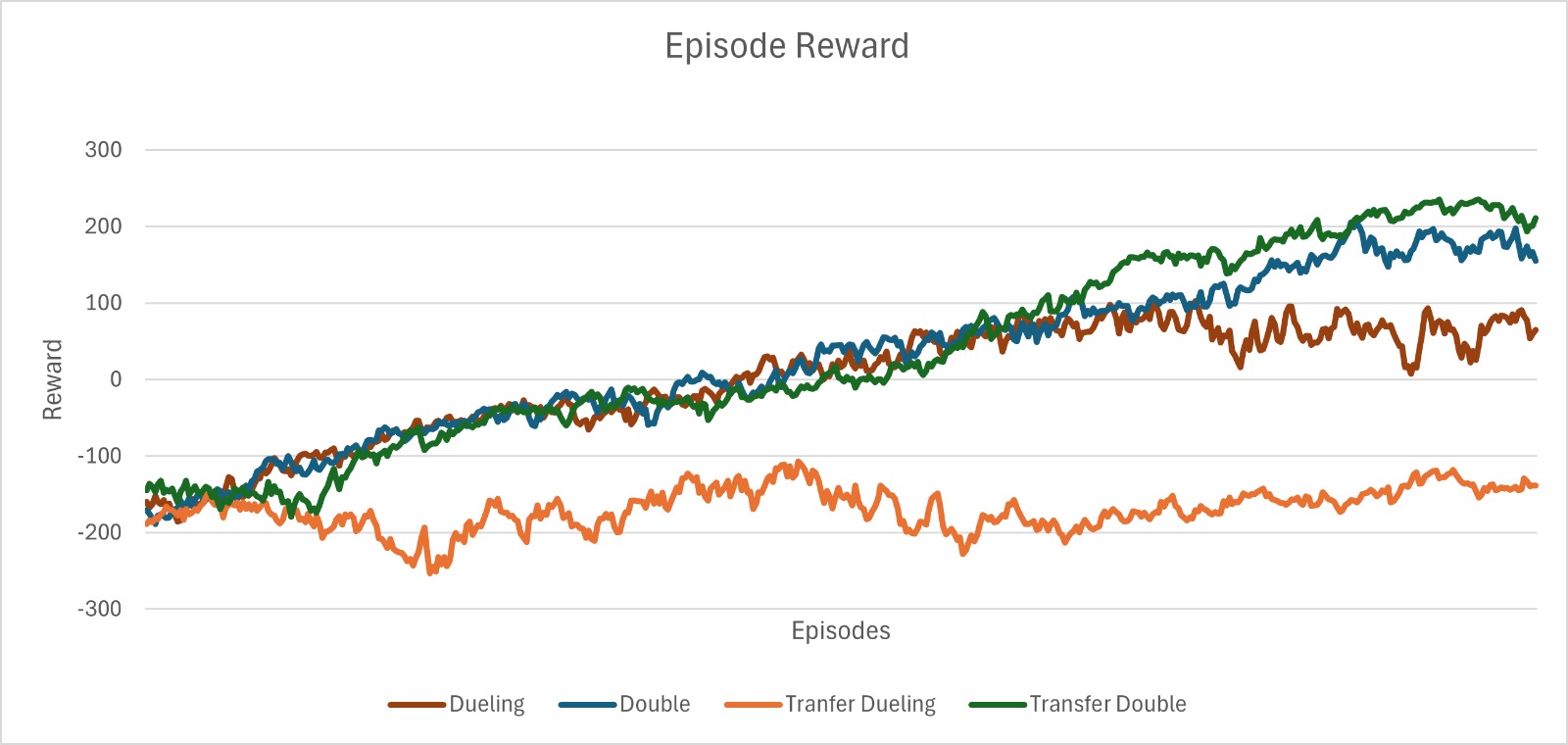}
\caption{Episode rewards: Transfer DDQN succeeds (+200) vs. Transfer Dueling fails (-150).}
\label{fig:episode_rewards}
\end{minipage}
\hfill
\begin{minipage}{0.32\textwidth}
\centering
\includegraphics[width=\textwidth]{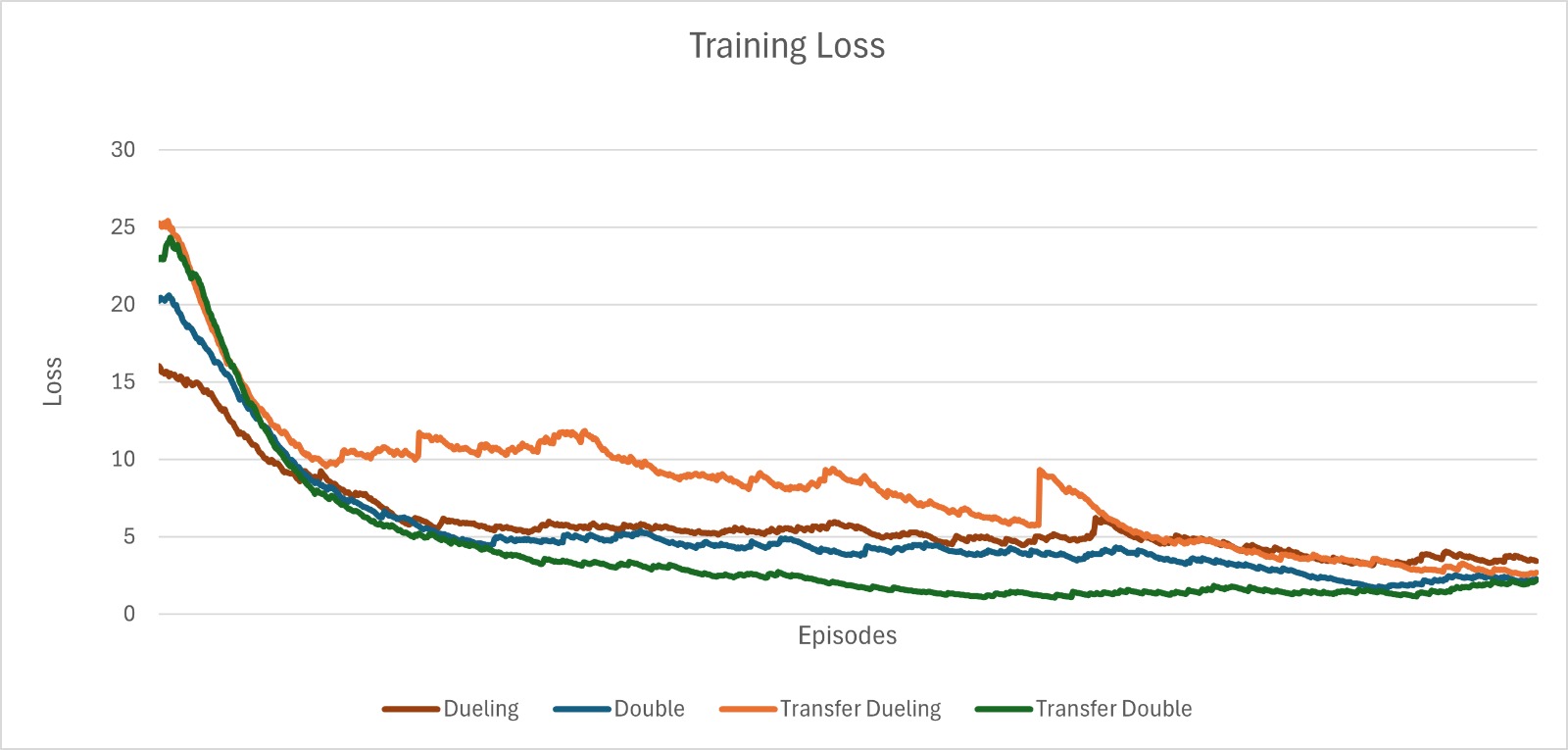}
\caption{Training loss: DDQN smooth ($<2$) vs. Dueling oscillatory (8-12).}
\label{fig:train_loss}
\end{minipage}
\end{figure*}

Both architectures achieved identical baseline performance on CartPole within 500 episodes, converging to near-optimal rewards (+200) with similar training dynamics. Transfer learning to LunarLander revealed pronounced differences. Transfer DDQN demonstrated robust positive transfer: validation rewards improved to -60, episode rewards reached +200-220, and training loss decreased smoothly to $\leq 2$. Transfer Dueling DQN exhibited consistent negative transfer: validation rewards stagnated at -370 (300-point deficit), episode rewards remained volatile around -150, and training loss plateaued at 8-12 with persistent oscillations. These patterns are consistent with DDQN's bias-reduction mechanism~\cite{2} may contribute to stable transferable representations, while Dueling DQN's advantage-value decomposition~\cite{15} introduces constraints preventing adaptation to sparse-reward environments~\cite{4,12,14,19}.

\textbf{Episode Rewards (Figure~\ref{fig:episode_rewards}).} CartPole baseline showed identical curves for both architectures (after smoothing, rewards increased from approximately -150 to +200). Transfer DDQN demonstrated robust positive transfer, reaching +200-220 by episode 500, matching baseline performance. Transfer Dueling DQN shows systematic performance degradation, remaining volatile around -150 with frequent crashes below -200. This 350-point gap suggests that Dueling's advantage stream~\cite{15}, optimized for CartPole's symmetric actions, provides architecture-specific mismatch for LunarLander's asymmetric dynamics~\cite{12,14}.

\textbf{Validation Rewards (Figure~\ref{fig:val_rewards}).} Baseline training on LunarLander showed identical trajectories (-200 to 0). Transfer DDQN maintained stability, improving from -200 to -60. Transfer Dueling DQN exhibited severe degradation, starting at -540 and reaching only -370. This 300-point gap suggests DDQN's bias-reduction~\cite{2} maintains stable policy evaluation under domain shift, while Dueling's decomposition~\cite{15} amplifies estimation errors when operating on mismatched features~\cite{4,19}.

\textbf{Training Loss (Figure~\ref{fig:train_loss}).} Baseline architectures showed similar reduction (15-25 to 2-3 by episode 500). Transfer DDQN maintained smooth monotonic reduction to <2, demonstrating stable TD error minimization~\cite{2}. Transfer Dueling DQN plateaued at 10-11 after episode 100 with persistent oscillations between 8-12, indicating unstable Q-value estimates. This contrast (loss $< 2$ versus $> 8$)
 directly explains the validation and episode reward gaps, confirming bias reduction produces more robust transfer than value decomposition~\cite{4,15,18}.





\section{Statistical Analysis}

\subsection{Descriptive Statistics}
To validate observed performance differences, we computed descriptive statistics across 5 random seeds for each architecture and setting. 

\begin{itemize}
    \item \textbf{DDQN Baseline} on LunarLander achieved a mean episode reward of $\mu = 171.01 \pm 37.53$, a mean validation reward of $\mu = 160.61 \pm 50.61$, and a training loss of $\mu = 2.09 \pm 0.77$. 
    \item \textbf{Dueling DQN Baseline} showed comparable mean performance, with an episode reward of $\mu = 62.94 \pm 148.63$ and validation reward of $\mu = 72.24 \pm 115.29$. However, it exhibited substantially higher variance in episode reward and validation reward ($\sigma = 148.63$ vs. $37.53$, a $3.96\times$ increase), indicating inherent instability even in single task settings.
\end{itemize}

In the \textbf{Transfer Learning} scenario, we observed dramatic divergence in performance:

\begin{itemize}
    \item \textbf{Transfer DDQN} achieved a significantly higher mean episode reward of $\mu = 212.18 \pm 17.13$, a mean validation reward of $\mu = 182.60 \pm 31.63$, and a much lower training loss of $\mu = 1.70 \pm 0.77$. This result demonstrates successful positive transfer with stable performance, outperforming the DDQN Baseline by 41 points in episode reward.
    \item \textbf{Transfer Dueling DQN} exhibited consistent negative transfer, with a mean episode reward of $\mu = -145.12 \pm 19.18$, validation reward of $\mu = -156.87 \pm 15.03$, and training loss of $\mu = 2.88 \pm 2.02$. The performance gap between \textbf{Transfer DDQN} and \textbf{Transfer Dueling} was substantial: a 357-point episode reward gap and a 339-point validation reward gap.
\end{itemize}

The statistical significance of these results was confirmed through a two-sample t-test, yielding a $p$-value of $< 0.001$. The Cohen’s d for episode rewards was extremely large ($d = 20.8$), reflecting near-complete separation between reward distributions rather than a precise estimate of practical magnitude. This unusually large effect size reflects near-complete separation between reward distributions and should be interpreted qualitatively rather than as a precise magnitude estimate. This analysis strongly suggests that \textbf{DDQN's bias-reduction mechanism}~\cite{2} is associated with robust, transferable representations, while \textbf{Dueling DQN's advantage-value decomposition}~\cite{15} exhibits instability under domain shift in our experiments~\cite{4,7}.Given consistent trends across random seeds and approximately normal reward distributions, a two-sample $t$-test was used.

\subsection{Inferential Statistics}

This section presents a detailed statistical analysis of transfer learning performance on the LunarLander environment. The objective of this analysis is to determine whether architectural differences between DDQN and Dueling DQN lead to systematically different outcomes under transfer, and how these outcomes compare to a baseline trained from scratch. Given the small number of random seeds ($n = 5$) and the known non-normality of reinforcement learning returns, the analysis emphasizes robust descriptive statistics and non-parametric inference.

All inferential analyses are conducted using \emph{per-seed aggregated validation rewards}. For each training configuration, validation rewards were aggregated over the final evaluation window, yielding one independent scalar value per seed. This aggregation preserves the independence assumptions required for statistical testing and avoids inflating sample size through temporally correlated episode-level measurements. Episode rewards and training loss are analyzed only descriptively and are used to contextualize learning dynamics and optimization behavior rather than to support hypothesis testing.

We first examine whether architectural design influences transfer performance when moving from CartPole to LunarLander. To this end, validation rewards obtained by the DDQN transfer configuration and the Dueling DQN transfer configuration were compared using a two-sided Mann--Whitney U test. The analysis reveals a statistically significant difference between the two configurations ($U = 25.0$, $p = 0.0079$). Across all five seeds, DDQN transfer consistently achieved positive validation rewards, whereas Dueling DQN transfer consistently resulted in negative validation rewards. This clear separation in reward distributions indicates that architectural inductive bias is strongly associated with transfer outcomes under severe domain shift in the examined setting. In particular, bias-reduction mechanisms in DDQN appear to provide greater robustness than advantage–value decomposition in Dueling DQN when representations learned in a simple environment are reused in a more complex one. This result provides a direct and affirmative answer to \textbf{RQ1}.

Next, we evaluate the effectiveness of transfer learning relative to a baseline trained from scratch on the target environment. When comparing DDQN trained from scratch on LunarLander with DDQN initialized via transfer, no statistically significant difference in validation reward is observed ($U = 8.0$, $p = 0.421$). This result suggests that, under the examined setup, transfer neither meaningfully improves nor degrades DDQN performance. In contrast, the comparison between the DDQN baseline and the Dueling DQN transfer configuration yields a statistically significant difference ($U = 25.0$, $p = 0.0079$), with the transferred Dueling DQN performing substantially worse than the baseline across all seeds. This constitutes strong evidence of negative transfer for the Dueling DQN architecture. Taken together, these comparisons demonstrate that transfer effectiveness is highly architecture-dependent, thereby addressing \textbf{RQ2}.

Stability under transfer is examined by analyzing the dispersion and variance of per-seed validation rewards. A Brown--Forsythe test, which is robust to non-normal distributions, does not detect a statistically significant difference in variance between DDQN transfer and Dueling DQN transfer configurations ($p = 0.121$). While this result does not provide conclusive statistical evidence of differing variance, it should be interpreted in light of the small sample size. Descriptively, DDQN transfer exhibits moderate dispersion around consistently positive rewards, whereas Dueling DQN transfer shows uniformly negative rewards with broader spread. These patterns suggest qualitative differences in stability that are not fully captured by variance-based hypothesis testing alone. Accordingly, \textbf{RQ3} is addressed primarily through combined inferential and descriptive evidence.

To further contextualize these findings, episode rewards and training loss trajectories are examined descriptively. DDQN transfer displays stable learning dynamics with sustained episode rewards and comparatively smooth loss curves, indicating consistent optimization behavior. In contrast, Dueling DQN transfer exhibits persistently negative episode rewards and more erratic loss trajectories, suggesting instability during value estimation and optimization. While these observations are not used for formal inference, they support the interpretation that optimization instability and representational mismatch contribute to failed transfer in the Dueling DQN configuration.

Overall, the statistical analysis demonstrates that architectural design strongly shapes transfer learning behavior under cross-environment domain shift. DDQN transfer maintains performance comparable to training from scratch, whereas Dueling DQN transfer exhibits systematic negative transfer. These findings highlight the fragility of naive representation reuse in value-based deep reinforcement learning and underscore the importance of architectural inductive bias in determining transfer robustness.

\section{Conclusion}

This work presented a controlled empirical comparison of Double Deep Q-Network (DDQN) and Dueling DQN architectures under cross environment transfer from CartPole to LunarLander. While both architectures achieved comparable baseline performance in the source environment, their transfer behavior diverged substantially. DDQN demonstrated robust positive transfer, maintaining stable learning dynamics and performance comparable to a baseline trained from scratch, whereas Dueling DQN exhibited severe negative transfer characterized by degraded rewards and unstable optimization.

Statistical analysis across multiple random seeds confirmed that these differences were significant, indicating that architectural inductive bias plays a critical role in determining transfer robustness in value-based deep reinforcement learning. In particular, DDQN’s bias-reduction mechanism appears more conducive to stable knowledge reuse under substantial domain shift than the advantage-value decomposition employed by Dueling DQN. These findings suggest that architectural choices optimized for single-task performance may not necessarily align with transfer learning objectives under cross-environment domain shift. This motivates future work on hybrid architectures and more robust transfer mechanisms.

\section{Limitations}

Due to computational constraints, all agents were trained for 500 episodes in the target domain, which was sufficient to observe transfer effects but insufficient for full convergence in LunarLander’s sparse-reward setting; extended training may reveal whether Dueling DQN eventually stabilizes or if its high variance persists~\cite{7,14}. In addition, the substantial structural disparity between CartPole (low-dimensional state, dense rewards) and LunarLander (higher-dimensional state, sparse rewards, complex dynamics) represents a challenging transfer scenario that may exceed the transferability of direct layer-wise reuse without additional mechanisms such as progressive unfreezing~\cite{8} or domain adaptation~\cite{19}. Methodologically, our study focused exclusively on direct weight transfer with early freezing, and alternative approaches including policy distillation~\cite{19}, meta-learned plasticity~\cite{9}, or task-specific adapter layers may mitigate the observed generalization difficulties. While multiple random seeds were used following best practices~\cite{7}, the modest number of trials and limited training duration constrain the statistical power of our findings. Finally, we evaluated canonical implementations of DDQN and Dueling DQN; hybrid architectures combining both innovations, such as those used in Rainbow~\cite{18}, or additional stabilization mechanisms like prioritized replay~\cite{5}, may exhibit different transfer characteristics. Accordingly, these results should be interpreted as empirical outcomes under a specific cross-environment transfer setting rather than universal properties of the architectures.

\section*{Acknowledgments}
AI Disclosure: The authors acknowledge the use of AI-assisted tools for language editing and clarity improvement. The scientific content, experimental design, and conclusions are solely those of the authors.

\vspace{12pt}


\begin{thebibliography}{00}



\bibitem{1} V. Mnih, K. Kavukcuoglu, D. Silver, A. A. Rusu, J. Veness,
M. G. Bellemare, A. Graves, M. Riedmiller, A. K. Fidjeland,
G. Ostrovski, et al., ``Human-level control through deep
reinforcement learning,'' \emph{Nature}, vol. 518, no. 7540,
pp. 529--533, Feb. 2015.

\bibitem{2} H. Van Hasselt, A. Guez, and D. Silver, ``Deep reinforcement
learning with double Q-learning,'' in \emph{Proc. AAAI Conf.
Artificial Intelligence (AAAI)}, 2016, pp. 2094--2100.

\bibitem{3} R. S. Sutton and A. G. Barto, \emph{Reinforcement Learning:
An Introduction}, 2nd ed. Cambridge, MA, USA: MIT Press, 2018.

\bibitem{4} M. E. Taylor and P. Stone, ``Transfer learning for reinforcement
learning domains: A survey,'' \emph{J. Mach. Learn. Res.},
vol. 10, pp. 1633--1685, Jul. 2009.

\bibitem{5} T. Schaul, J. Quan, I. Antonoglou, and D. Silver,
``Prioritized experience replay,'' in \emph{Proc. Int. Conf.
Learning Representations (ICLR)}, 2016.

\bibitem{6} K. Arulkumaran, M. P. Deisenroth, M. Brundage, and
A. A. Bharath, ``A brief survey of deep reinforcement learning,''
\emph{IEEE Signal Process. Mag.}, vol. 34, no. 6,
pp. 26--38, Nov. 2017.

\bibitem{7} P. Henderson, R. Islam, P. Bachman, J. Pineau,
D. Precup, and D. Meger, ``Deep reinforcement learning that matters,''
in \emph{Proc. AAAI Conf. Artificial Intelligence (AAAI)},
pp. 3207--3214, 2018.

\bibitem{8} A. A. Rusu, N. C. Rabinowitz, G. Desjardins, H. Soyer,
J. Kirkpatrick, K. Kavukcuoglu, R. Pascanu, and R. Hadsell,
``Progressive neural networks,'' \emph{arXiv preprint}
arXiv:1606.04671, Jun. 2016.

\bibitem{9} C. Finn, P. Abbeel, and S. Levine,
``Model-agnostic meta-learning for fast adaptation of deep networks,''
in \emph{Proc. 34th Int. Conf. Machine Learning (ICML)},
pp. 1126--1135, 2017.

\bibitem{10} J. Schulman, F. Wolski, P. Dhariwal, A. Radford,
and O. Klimov, ``Proximal policy optimization algorithms,''
\emph{arXiv preprint} arXiv:1707.06347, Jul. 2017.

\bibitem{11} B. Ben, ``Solving Gymnasium's Lunar Lander with deep
Q-learning (DQN),'' Finding Theta, 2023. [Online]. Available:
https://www.findingtheta.com/blog/solving-gymnasiums-lunar-lander-with-deep-q-learning-dqn

\bibitem{12} A. M. Elsaid and I. A. Aly, ``Evaluating reinforcement learning
algorithms for LunarLander-v2: A comparative analysis of DQN,
DDQN, DDPG, and PPO,'' \emph{ResearchGate}, Feb. 2024.
[Online]. Available:
https://www.researchgate.net/publication/388662081

\bibitem{13} K. He, X. Zhang, S. Ren, and J. Sun,
``Deep residual learning for image recognition,''
\emph{arXiv preprint} arXiv:1512.03385, Dec. 2015.

\bibitem{14} S. Kancharla,
``Deep Q-learning optimal parameter search: Lunar Lander simulation,''
\emph{arXiv preprint} arXiv:2209.07809, Sep. 2022.

\bibitem{15} Z. Wang, T. Schaul, M. Hessel, H. van Hasselt,
M. Lanctot, and N. de Freitas,
``Dueling network architectures for deep reinforcement learning,''
\emph{arXiv preprint} arXiv:1511.06581, Apr. 2016.

\bibitem{16} S. Mishra and A. Arora,
``Double deep Q network with Huber reward function for Cart-Pole
balancing problem,'' \emph{arXiv preprint} arXiv:2006.04938,
Jun. 2020.

\bibitem{17} K. Saadat and R. Zhao,
``Enhancing two-player performance through single-player knowledge
transfer: An empirical study on Atari 2600 games,''
\emph{arXiv preprint} arXiv:2410.16653, Oct. 2024.

\bibitem{18} M. Hessel, J. Modayil, H. van Hasselt, T. Schaul,
G. Ostrovski, W. Dabney, D. Horgan, B. Piot, M. G. Azar,
and D. Silver, ``Rainbow: Combining improvements in deep
reinforcement learning,'' \emph{arXiv preprint}
arXiv:1710.02298, Oct. 2017.

\bibitem{19} Z. Zhu, K. Lin, A. K. Jain, and J. Zhou,
``Transfer learning in deep reinforcement learning: A survey,''
\emph{IEEE Trans. Pattern Anal. Mach. Intell.}, 2023,
doi: 10.1109/TPAMI.2023.3292075.





\end{thebibliography}
\end{document}